\documentclass[a4paper, 10pt, conference]{ieeeconf}      % Use this line for a4 paper

\IEEEoverridecommandlockouts                              % This command is only needed if 
\usepackage{epsfig} % for postscript graphics files
\usepackage{mathptmx} % assumes new font selection scheme installed
\usepackage{comment}
\usepackage{graphicx} % for pdf, bitmapped graphics files
\usepackage{subcaption}
\usepackage{caption}

\makeatletter
\let\NAT@parse\undefined
\makeatother

\usepackage{amsmath} % assumes amsmath package installed

\usepackage{amssymb}  % assumes amsmath package installed
\usepackage{pgfplots}

\usepackage{algorithm}
\usepackage{algpseudocode}
\usepackage{tabularx}
\usepackage{url}
\usepackage{amssymb}

\usepackage{cleveref}
\usepackage{cite}
\usepackage{etoolbox}
\makeatletter
\patchcmd{\@makecaption}
  {\scshape}
  {}
  {}
  {}
\makeatother

\usepackage[letterpaper,bindingoffset=0.2in,%
left=0.75in,right=0.75in,top=0.75in,bottom=0.75in,%
footskip=.25in]{geometry}

\usepackage{accents}

\usepackage{pgfplots}
\usepackage{lipsum}
\usepackage{xcolor}
\usepackage{pgfplots}

\usepackage{multirow}

\title{

\LARGE \bf A Neuro-Symbolic Approach for Enhanced Human Motion Prediction 

\author{Sariah Mghames$^{1}$, Luca Castri$^1$, Marc Hanheide$^1$, Nicola Bellotto$^{1,2}$
\thanks{\noindent\textsuperscript{1}School of Computer Science, University of Lincoln, UK.\newline
\textsuperscript{2}Dept. of Information Engineering, University of Padua, Italy.\newline
This project has received funding from the EU's Horizon 2020 Research and Innovation programme under grant agreement No 101017274}
}
}
\begin{document}

\maketitle

\begin{abstract}
Reasoning on the context of human beings is crucial for many real-world applications especially for those deploying autonomous systems (e.g. robots). In this paper, we present a new approach for context reasoning to further advance the field of human motion prediction. We therefore propose a neuro-symbolic approach for human motion prediction (NeuroSyM), which weights differently the interactions in the neighbourhood by leveraging an intuitive technique for spatial representation called Qualitative Trajectory Calculus (QTC).
 The proposed approach is experimentally tested on medium and long term time horizons using two architectures from the state of art, one of which is  a baseline for human motion prediction and the other is a baseline for generic multivariate time-series prediction. Six datasets of challenging crowded scenarios, collected from both fixed and mobile cameras, were used for testing. 
Experimental results show that the NeuroSyM approach outperforms in most cases the baseline architectures in terms of prediction accuracy. 

\end{abstract} 

%\lipsum[1]

\section{Introduction}
Human motion prediction has been the area of focus of many researchers to date, ranging from single human motion prediction (i.e. with no context) to the most developed frameworks in context-aware (dynamic and static context) human motion prediction. The importance given to this area of study  traces back to the crucial impact it has on many real-world applications including but limited to video surveillance, anomaly detection, action and intention recognition, autonomous driving, and robot navigation. While many studies on human motion prediction have been relying on datasets collected from a fixed camera to enhance the accuracy and time complexity of their frameworks, very few have studied the field from a mobile camera perspective where the problem becomes more challenging with restriction on the global observability of the scene context and interactions. Hence, we focus in this work on studying the field from both fixed and mobile camera perspective targeting the autonomous systems (e.g. robotic) application.

Reasoning on the context (e.g multi-agents interactions, static key objects) of humans is crucial primarily for safe autonomous systems navigation where human-robot co-existence, for example, is increasingly taking part in domestic, healthcare, warehouse, and transportation domains. A robot tasked to deliver an order to a table in a restaurant, bring a medicine to a patient in a hospital, or clean a road side-walks, needs to update on-the-fly its internal state representation of dynamic agents in the scene and, therefore, update its target plan.

\begin{figure}[t]\centering
\includegraphics[scale=0.25]{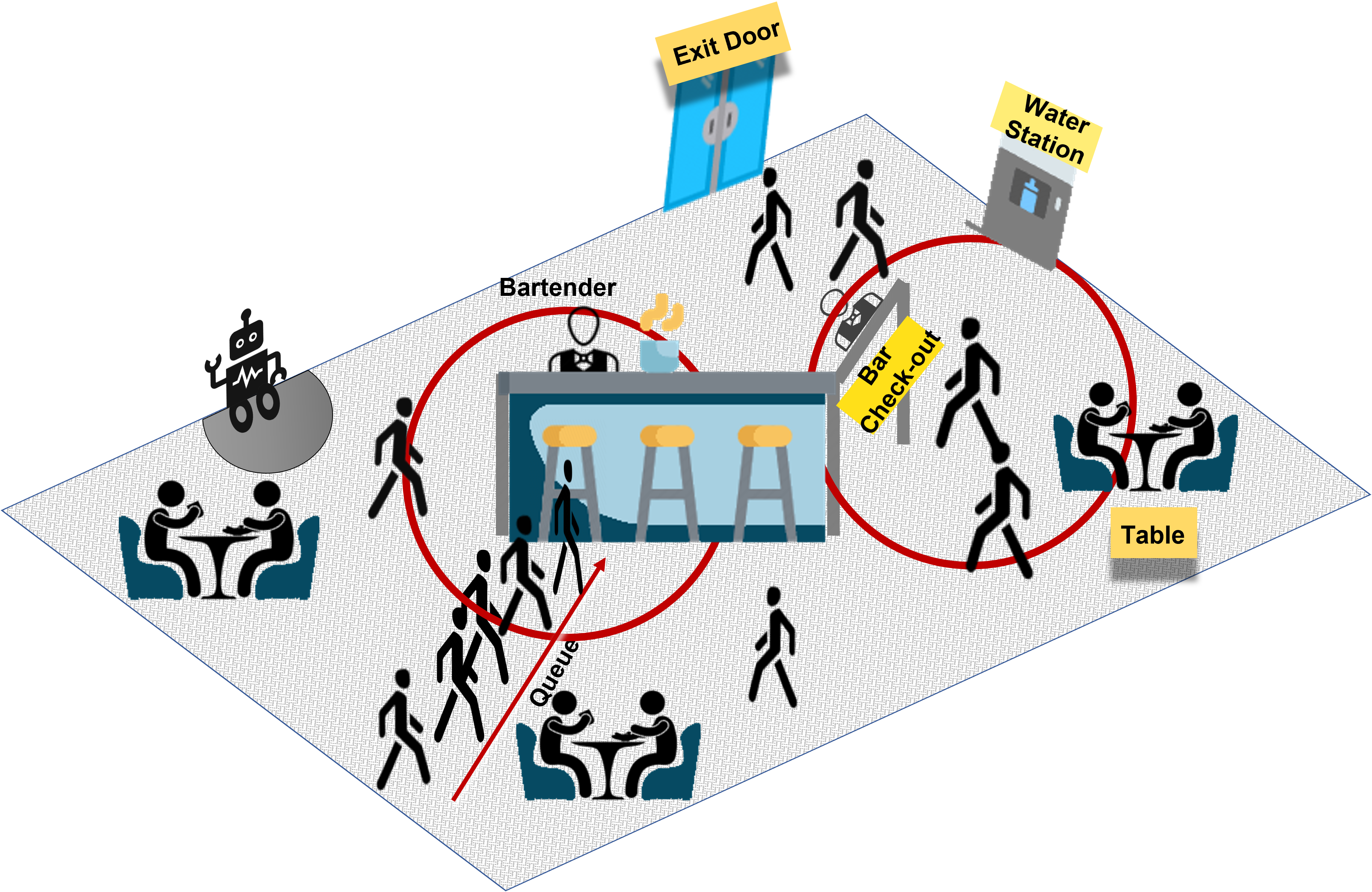}
\caption{Conceptual illustration of the social cafe-bar scenario used for reasoning on the context and interactions of humans in dense environments. In the area of context-aware human motion prediction, a research gap remains for embedding all the neighbourhood interactions in the learning process without reasoning on the ones that are more or less stable (i.e. reliable) than others, and hence can be more or less important in affecting the future states of a single agent.} \label{fig:intro}
%This can be realised by looking at the queue and the table, where not every relative motion in the neighborhood influences the future states of a single human. }
\end{figure}

In addition to safety in navigation, reasoning on the motion of multi-agents presents also the advantage of implicit intent communication.
For example, a social robot detecting a conversational group in the environment and predicting that the group will hold on its current interaction for a while, makes sure to not unnecessarily interfere with the group. On the other hand, if the robot predicts that a person is coming towards itself, e.g to handle a box in a warehouse, it can select an action that prioritizes the responsiveness to the human's intent. 
%From here arises the importance of reasoning on the context for human motion prediction and for implicit multi-agents intent communication. 

Context reasoning is presented over the literature in the form of human-human and human-objects interactions reasoning. The authors  in~\cite{chen2019crowd} have jointly modeled human-robot and human-human interactions in a deep reinforcement learning framework to drive robot navigation. 
In~\cite{pokle2019deep}, instead, the authors learn an optimal local trajectory from a global plan by fusing human trajectories, Lidar features, global path and odometry features in an attention layer. 
Context-awareness methods have also been proposed to deal with the challenges faced by the long-term prediction of single human motion~\cite{Alahi2016,lisotto2019social,gupta2018social,mangalam2020not,liang2019peeking,Cao2020}. In the previous works, interactions are processed either in a grid-based pooling approach or in a global pooling mechanism to deal with the problem of dynamic neighbourhood size. 

Though context-awareness has proven better accuracy in predicting human motion, one problem remains unnoticed. Indeed, when interactions are defined spatially and hence retrieved from the relative motion between pairs of agents, not all neighbourhood interactions are of equal reliability and hence of equal importance to the prediction of future states of a single human being, as can be the case for the spatial interactions interconnecting the queue line and the table in Fig.~\ref{fig:intro}.
%For example, when people wait in a queue at the cafe-bar (as in Fig.~\ref{fig:intro}) the interaction between the tale and leading persons may not affect the future states of the leading one. 
In this work, we address the problem of context-aware human motion prediction by injecting a-priori information on the interactions in a neuro-symbolic approach.
Among spatial interaction representations, the qualitative approach presents an intuitive way for interaction description. Qualitative spatial interactions are defined as symbolic representations of interactions between a pair of agents in the spatial domain, i.e. 2D navigation. One way to model qualitative spatial interactions in multi-agent scenarios is by using the qualitative trajectory calculus (QTC) \cite{delafontaine2012qualitative,bellotto2013qualitative}. QTC-based models of moving agent pairs can be described by different combinations of QTC symbols that represent spatial relations between pairs of interacting agents, like relative distance (i.e moving towards/away), velocity (i.e moving faster/slower), and orientation (i.e. moving to left/right).

The contribution of this paper is therefore three-fold: (i)~proposing a novel neuro-symbolic approach for enhancing human motion prediction (denoted NeuroSyM) using a-priori information on the spatial interactions between couple of agents, (ii) experimentally evaluating the proposed framework on two architectures from the state of art, one of which is a baseline for human motion prediction and the other is a baseline for generic multivariate time-series prediction, and on different datasets  collected from both fixed and mobile camera perspective, (iii) releasing the source code as a \textit{Github} repository\footnote{https://github.com/sariahmghames/NeuroSyM-prediction} with some qualitative results to help in the testing and integration of NeuroSyM on other baselines for human motion prediction.

The remainder of the paper is as follows: Sec.~\ref{sec:lit} presents an overview of the related works; Sec.~\ref{sec:appr} explains the approach adopted to reason on the context of humans in dense scenes; Sec.~\ref{sec:exper} illustrates and discusses the results from experiments conducted on open-source datasets for human motion prediction and social navigation; finally, Sec.~\ref{sec:conc} concludes by summarising the main outcomes and suggesting future research work.

 \label{sec:intro}

\section{Related Works} \label{sec:lit}

%\textbf{Human-Objects interactions modeling:}
\textbf{Context-aware human motion prediction:} 
The state of the art shows extensive works in the area of context-aware human motion prediction. Among those works, some incorporate spatio-temporal dependencies~\cite{tao2020dynamic,yu2020spatio,huang2019stgat} of interactions and others are limited to only spatial~\cite{Alahi2016,lisotto2019social,gupta2018social,mangalam2020not,liang2019peeking,Cao2020,lee2017desire,bisagno2021embedding}. Another sub-category differentiates related works into those dealing with both dynamic and static context~\cite{tao2020dynamic,lisotto2019social,liang2019peeking,lee2017desire,bisagno2021embedding}, those neglecting the dynamic context~\cite{Cao2020}, while others focus more on the dynamic context of interactions only~\cite{Alahi2016,gupta2018social,mangalam2020not,yu2020spatio,huang2019stgat}.

The Dynamic and Static Context-aware Motion Predictor (DSCMP) in~\cite{tao2020dynamic} integrates dynamic interactions between agents in a Social-aware Context  Module (SCM), whereas the static context is incorporated in a latent space with a semantic scene mapping. The two most common baseline architectures used over the literature for human motion prediction are the Social-LSTM (S-LSTM)~\cite{Alahi2016} and the Social Generative Adversarial Networks (SGAN) ~\cite{gupta2018social}. They use a spatially-aware pooling mechanism for incorporating the hidden states of proximal dynamic agents as a way to overcome the problem of variable and (potentially) large number of people in a scene. The SGAN, however, has outperformed S-LSTM in terms of both accuracy and time complexity by avoiding the grid-based pooling mechanism technique. In parallel, SGAN outperformed Stgat~\cite{huang2019stgat} with time complexity and parameters consumption.

In this work, the fundamental SGAN architecture from the literature is used to evaluate our NeuroSyM approach for motion prediction, leaving room for potential integration of other architectures with static context awareness (e.g. the image-driven static context of~\cite{tao2020dynamic}). Here we consider only raw trajectories (or metric coordinates) of the context (dynamic and/or static) as possible input to the deployed model architectures.

\textbf{Human-human interactions modeling:}
The methods for interactions modeling with nearby dynamic agents can be classified into two types of problem: (a) one-to-one modeling, and (b) crowd modeling~\cite{Ijaz2015,hedayati2020reform,thompson2021conversational}.
A one-to-one interaction modeling was presented in the literature in the form of quantitative or qualitative representations. Quantitative representation of interactions leverages a multi-layer perceptron to embed relative pose (positions or velocities) between pairs of agents as in~\cite{chen2019crowd,gupta2018social,Alahi2016}. While qualitative representation of interactions was used in~\cite{dondrup2014probabilistic} and~\cite{hanheide2012analysis} to model human-robot spatial interactions using QTC. In~\cite{dondrup2014probabilistic} the use of qualitative rather than quantitative representations for analysing human-robot spatial interactions (HRSI) was motivated by the need of a more intuitive understanding of the observed interactions. In~\cite{dondrup2016qualitative} and~\cite{bellotto2012robot} similar models are used to implement human-aware robot navigation strategies.
The prediction of interactions in \cite{dondrup2016qualitative} is based on a Bayesian temporal model limited to single human-robot pairs, without considering nearby static or dynamic objects, which limits the prediction performance.

%Spatial interactions between pairs of agents, i.e. human-human or human-robot, have been usually modeled using a probabilistic approach~\cite{dondrup2014probabilistic}, e.g. by encoding a sequence of QTC states in a Markov Chain representation.

%On the other hand, crowd modeling was discussed in~\cite{Ijaz2015}, where the major existing hybrid techniques were surveyed. Hybrid crowd techniques are brought forward to overcome some limitations of classical methods (e.g high computation cost). They successfully handle large crowd size while considering heterogeneous individual behaviors and taking into account three major factors for crowd modeling: (a) crowd size, (b) need for implementation of individual behaviors, (c) computation cost. 

%Following the crowd analysis, F-formations modeling and detection has been addressed recently in~\cite{hedayati2020reform} and~\cite{thompson2021conversational}. 

%In~\cite{hedayati2020reform}, the authors deconstructed a social scene into pairwise data points, then they used feature-based (distance and effort angle) classification to distinguish F-formations. The work in~\cite{thompson2021conversational}, instead, proposes a Graph Neural Network~(GNN) to predict pairwise affinities for an interaction graph, and from them the likelihood that two people are part of an F-formation. Affinities are then used to cluster people into conversational groups.

In our study of single human motion, we rely on one-to-one (i.e. pairwise) interactions modeling due to the different nature of interactions that may occur in the neighbourhood of a single agent. Hence, we build on previous works from qualitative representation of interactions~\cite{dondrup2014probabilistic} to weight the quantitative embedding of neighbourhood interactions.
%taking inspiration from hybrid approaches of crowd modeling for crowdy scenes.

\section{NeuroSyM Prediction Approach} \label{sec:appr}
 \subsection{Problem Definition}

While most, if not all, works in the area of context-aware human motion prediction embeds equally all kind of interactions in the pre-defined neighbourhood size of a single agent, in this work we formulate the problem of context-aware human motion prediction in terms of weighted interactions embedding between pairs of agents. We show that a-priori information on the kind of interactions helps the network to predict motion with a better accuracy.
In the following, we present the formulation of spatial interactions which will be used later on to label (or weight) the interactions as a symbolic reasoning called by the neural model.

\subsection{Spatial Interactions: a Qualitative Formulation} \label{sec:qtcform}

\begin{figure}\centering
\includegraphics[trim={10cm 6cm 12cm 3cm},clip, scale=0.27]{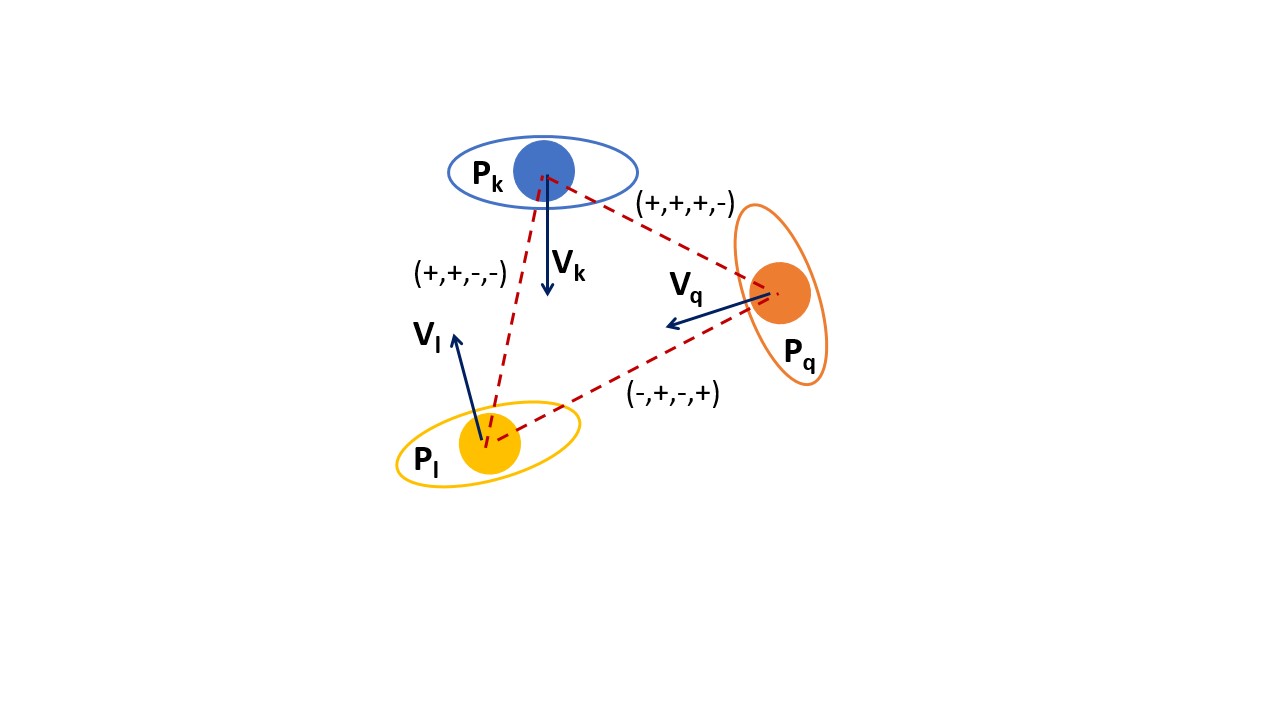}
\caption{A case of $QTC_{C_1}$ representation of interactions between three body points $P_k$, $P_l$, and $P_q$.}\label{fig:qtcc}
\end{figure}

A qualitative spatial interaction is defined by a vector of $m$ QTC relations~\cite{delafontaine2012qualitative}, which consist of qualitative symbols ($q_i$, $i \in \mathbb{Z}$) in the domain $U=\{-, 0, +\}$. We can distinguish between four types of QTC: (a) $QTC_{B}$ basic, (b) $QTC_C$ double-cross, (c) $QTC_{N}$ network, and (d) $QTC_{S}$ shape. Here, we focus on the use of $QTC_{C}$, since it better represents the dynamics of the agents in our application scenario.
 Two types of $QTC_{C}$ exist in the literature: $QTC_{C_1}$, with four symbols $\{q_1, \, q_2, \, q_3, \, q_4\}$, and $QTC_{C_2}$, with six symbols $\{q_1, \, q_2, \, q_3, \, q_4, \, q_5, \, q_6\}$. The symbols $q_1$ and $q_2$ represent the towards/away (relative) motion between a pair of agents; $q_3$ and $q_4$ represent the left/right relation; $q_5$ indicates the relative speed, faster or slower; finally, $q_6$ depends on the (absolute) angle with respect to the reference line joining a pair of agents. The $QTC_{C_1}$ type is illustrated in Fig.~\ref{fig:qtcc} for a case of interaction between three body points. Given the time series of two moving points, $P_k$ and $P_l$, the qualitative interaction between them is expressed by the symbols $q_i$ as follows: 

\begin{small}
\centering
\begin{align*}
(q_1)  ~ &- : d(P_k|t^-, P_l|t) > d(P_k|t, P_l|t) \\
&0: d(P_k|t^-, P_l|t) = d(P_k|t, P_l|t) \\
 &+: d(P_k|t^-, P_l|t) < d(P_k|t, P_l|t) \\
(q_2) ~ &\text{same as $q_1$, but swapping $P_k$ and $P_l$} \\
(q_3) ~ &-: \|\vec{P_k^{t^+}P_k^t} \wedge \vec{P_l^tP_k^t}\| < 0 \\
 &0: \|\vec{P_k^{t^+}P_k^t} \wedge \vec{P_l^tP_k^t}\| = 0 \\
&+: \text{all other cases} \\
(q_4) ~ &\text{same as $q_3$, but swapping $P_k$ and $P_l$} \\
(q_5) ~ &-: \|\vec{V_k^t}\| < \|\vec{V_l^t}\| \\
&0: \|\vec{V_k^t}\| = \|\vec{V_l^t}\| \\
 &+: \text{all other cases} \\
(q_6) ~ &-: \theta(\vec{V_k^t}, \vec{P_kP_l}^t) < \theta(\vec{V_l^t}, \vec{P_lP_k}^t) \\
 &0: \theta(\vec{V_k^t}, \vec{P_kP_l}^t) = \theta(\vec{V_l^t}, \vec{P_lP_k}^t) \\
 &+: \text{all other cases.} 
\end{align*}
\end{small}

\noindent where $d(.)$ is the euclidean distance between two positions, $V(.)$ the velocity vector of a body point, $\theta(.)$ is the absolute angle between two vectors, and $\wedge$ is the cross-product notation between two vectors. 
In this paper, we propose a neuro-symbolic approach for motion prediction (NeuroSyM) that can be implemented to every related work in the field to enhance the accuracy of the motion. In order to narrow down the study, we take advantage of $QTC_{C_1}$ to label the interactions as described in the following section. We leave therefore the investigation into the additional information provided by $QTC_{C_2}$ to our future work.

\subsection{Data Labeling} \label{sec:data_labeling}
\begin{figure}\centering
\includegraphics[trim={12cm 6cm 12cm 3cm},clip, scale=0.35]{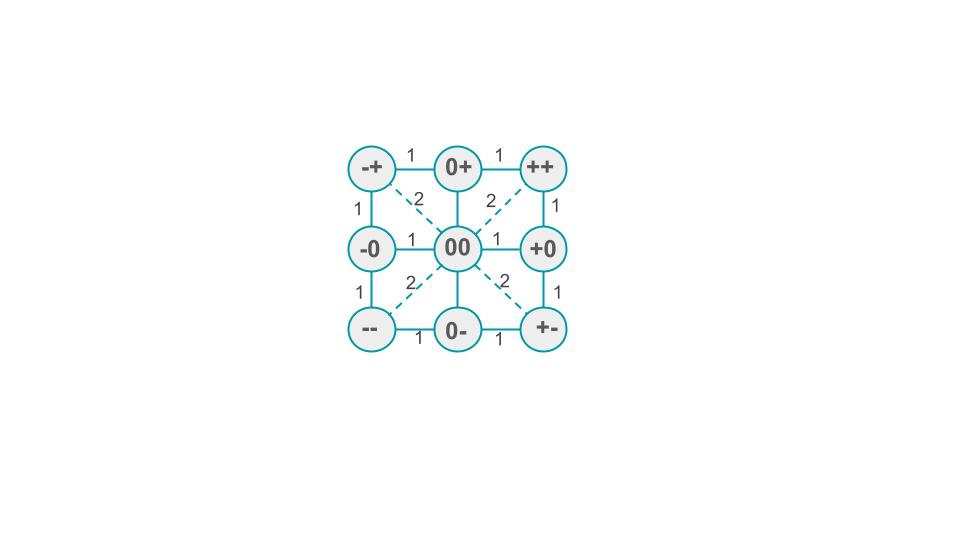}
\includegraphics[trim={10.5cm 4.5cm 11.5cm 3cm},clip, scale=0.35]{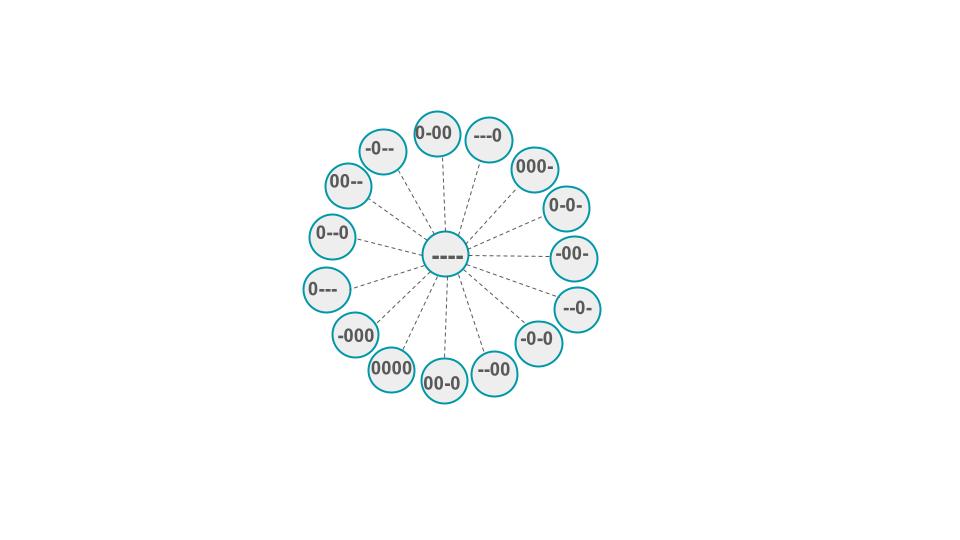}
\caption{(left) The complete CND for $QTC_{B1}$ in n-dimensional space. The $QTC_{B1}$ has only $q_1$ and $q_2$ symbols. Straight edges represent a conceptual distance of 1 while dashed edges represent a conceptual distance of 2. (right) one part of the $QTC_{C1}$ illustrating the possible transitions of the QTC relation $\{-, -, -, -\}$, resulting in 15 possible transitions and therefore in $\alpha_{cnd}$ of 0.067 weighting that same QTC relation at the next time step.}\label{fig:qtccnd}
\end{figure}

We leverage our labeling technique for pairwise spatial interactions on the concept of Conceptual Neighbourhood Diagram (CND) presented in~\cite{van2005conceptual} and in the original work of qualitative spatial interactions in~\cite{delafontaine2012qualitative}. 
As per~\cite{delafontaine2012qualitative}, the construction of a CND (as in Fig.~\ref{fig:qtccnd}) for QTC is based on the notion of conceptual distance ($\mathbf{d}$), which is used to define the closeness of two QTC states at time t and t', respectively, and can be calculated as follows:
\begin{equation}\label{eq:ddqtc}
\mathbf{d}_{QTC^t}^{QTC^{t'}} = \sum_{q_i} \mid  q_i^{QTC^t} - q_i^{QTC^{t'}} \mid, 
\end{equation}

\noindent where, for practical reasons, the symbols ``+'' and ``-'' are associated to the numerical values ``+1'' and ``-1'', as in~\cite{delafontaine2012qualitative}. In Fig.~\ref{fig:qtccnd} (left), for each link (i.e. edge) between conceptual neighbours (the nodes) the conceptual distance between the adjacent relations is indicated. 
In a CND, and due to the laws of continuity, the conceptual neighbours of each particular relation constitute only a subset of the base relations. For example, $QTC_{C_1}$ has 81 basic states or relations (each symbol $q_i$ has 3 different possible types of transitions from domain $U$) but the conceptual neighbours of $\{-, -, -, -\}$ relation as illustrated in Fig.~\ref{fig:qtccnd} (right) reduce from 80 to 15 for the following reasons~\cite{van2005conceptual}:
\begin{itemize}
    \item Transition from ``+" to ``-" (and vice versa) is impossible without passing through 0, hence transition from $\{-, -, +, +\}$ to $\{-, +, +, +\}$ is impossible without passing through $\{-, 0, +, +\}$.
    \item ``0" Dominates ``+" and ``-", hence a transition from $\{+, -, -, 0\}$ to $\{+, -, 0,+\}$ is impossible without passing through $\{+, -, -, +\}$ or $\{+, -, 0, 0\}$.
    \item The combination of both former rules.
\end{itemize}

For the sake of labeling, we omit the need for information on the conceptual distance between states, and we focus on the possible transitional states for each QTC relation given the state at time t. The CND for $QTC_{C_1}$ is not completely shown in Fig.~\ref{fig:qtccnd} (right) as it is too complex to visualise on a two-dimensional medium. The label ($\alpha_{cnd}$) for each of the 81 states of a $QTC_{C1}$ type of qualitative calculus is formulated as follows:

\begin{equation} \label{eq:labeling}
    \alpha_{cnd} = \Pr(QTC^{t'} | QTC^t) = \frac{1}{N_{Tr}}
\end{equation}
where $N_{Tr}$ represents the number of transitional states. $\alpha_{cnd}$ 
 represents the level of stability or reliability of a transitional state. The higher the number of possible transitional states, the lower the likelihood to transition into a single state and vice versa. In Fig.~\ref{fig:qtccnd}, the likelihood to transition from   $\{-, -, -, -\}$ into $\{0, 0, 0, 0\}$ is 0.067, however the likelihood increases to 0.2 if the 15 possible transitional states reduces to 5, rendering the $\{0, 0, 0, 0\}$ state more reliable in the learning process. Given an interaction at time t, we associate its label to the interaction (observed or predicted) at $t+1$. In most related works, an interaction between agents A and B is calculated as an embedding of the relative pose between them, as follows:

\begin{equation}\label{eq:inter}
     Inter_{AB} = \emph{Dense} (X_B-X_A)
\end{equation}
where $Dense()$ is the embedding layer. Given the pose X of each agent at time t, a QTC state can be formulated and the  corresponding label ``$\alpha_{cnd}$'' will be loaded from a dictionary. Hence, the symbolic reasoning transforms Eq.~\ref{eq:inter} into the form of $\alpha_{cnd}~Inter_{AB}$.
From a practical point of view, the symbolic knowledge of interactions between  two moving body-points can be readily exploited by any neural architecture for context-aware motion prediction, since the CND dictionary (associating QTC states with their corresponding $\alpha_{cnd}$), generated for a specific QTC configuration, remains the same regardless of the data distribution domain.

\section{Experiments} \label{sec:exper}

\begin{figure}
     \centering
     \begin{subfigure}[b]{0.38\columnwidth}
        \centering
         \includegraphics[width=\textwidth]{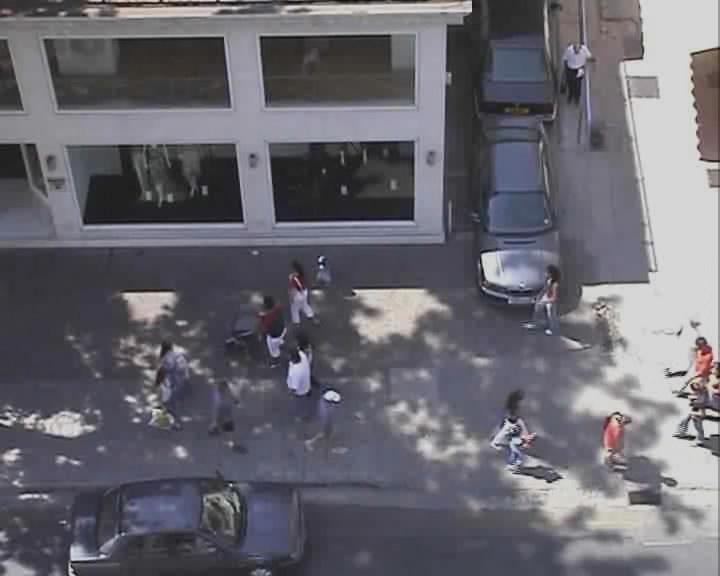}
         \caption{Zara dataset}
         \label{fig:zara}
     \end{subfigure}
     \hspace{5pt}
     \begin{subfigure}[b]{0.38\columnwidth}
         \centering
         \includegraphics[width=\textwidth]{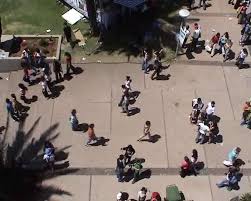}
         \caption{University dataset}
         \label{fig:univ}
     \end{subfigure}
     \begin{subfigure}[b]{\columnwidth}
         \centering
         \includegraphics[width=\textwidth]{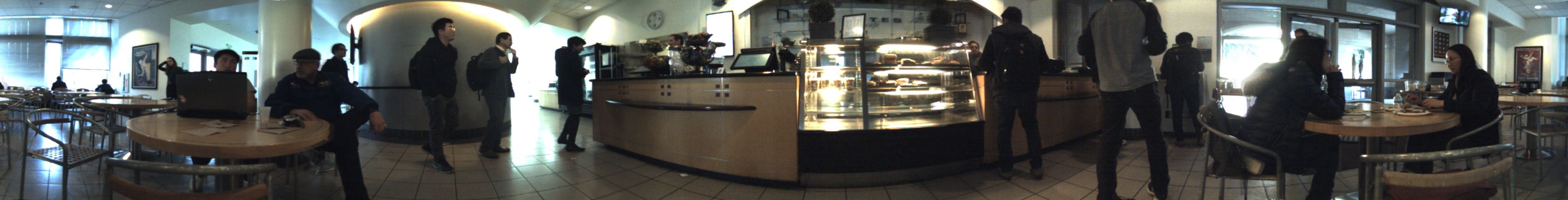}
         \caption{JackRabbot cafe \textit{bytes-cafe-2019-02-07\_0} scene}
         \label{fig:cafe}
     \end{subfigure}
        \caption{Examples from UCY and JackRabbot datasets.}
        \label{fig:dataset}
\end{figure}

As anticipated in Sec.~\ref{sec:lit}, we evaluated our approach for enhancing human motion prediction on two architectures from the state of the art, using raw trajectories as input. The first one is a well-known baseline architecture, the socially-acceptable trajectories with  generative adversarial networks (SGAN \cite{gupta2018social}), used in the literature to enhance the accuracy and speed of human motion prediction in crowds. It relies on datasets collected from a fixed top-down camera in public spaces, capturing the entire scene dynamics (ETH dataset~\cite{ess2007depth} -- sequences ETH and Hotel), and on datasets collected from a mobile stereo rig mounted on a car (UCY dataset~\cite{lerner2007crowds} -- sequences Zara01-02 and Univ). In order to generalise our evaluation to robotics application, we chose another well-known dataset, the JackRabbot (JRDB~\cite{martin2019jrdb}), which provides multi-sensor data of human behaviours from a mobile robot perspective in populated indoor and outdoor environments (Fig.~\ref{fig:cafe}). JackRabbot was never exploited in the literature for human motion prediction, although 
%and in addition can't be deployed on architecture like SGAN whose main contribution was the embodiment of global interactions in the scene. 
it clearly benefits applications of social robot navigation, where local interactions can be extracted from the on-board 360$^\circ$ Lidar (Velodyne) and Fisheye camera sensors. To this end, we chose to use JackRabbot on a generic network architecture for time series prediction and where the following features can be incorporated: (a) the ability to integrate a dynamic context; (b) the ability to integrate key static objects of potential interactions (e.g. door, table, bar), differently from S-LSTM and SGAN; (c) the ability to test our neuro-symbolic approach on prediction architectures that, instead of using a pooling mechanism to overcome the size problem of dynamic input series (representing the neighbourhood in social scenarios), weights every single input (i.e. neighbour) by giving special attention to each one separately. One of the recent  architectures that satisfy the last features is the dual-stage attention mechanism (DA-RNN) developed for time-series forecasting in~\cite{qin2017dual}. 

\subsection{Neuro-Symbolic SGAN} \label{sec:sgan}

\textit{\textbf{NeuroSyM SGAN Architecture:}} The original SGAN architecture (SGAN-20VP-20 in~\cite{gupta2018social}) has proven good performance in terms of accuracy, collision avoidance, and time complexity with respect to its precedent baselines, as the social LSTM~\cite{Alahi2016}. The core of SGAN is a generator and a discriminator trained adversarially. The generator model \textit{G} has the role of generating candidate trajectories, while the discriminator model \textit{D} estimates the probability that a sample comes from the training data (i.e. real) rather than from the generator output samples. 
The generator consists of an encoder and a decoder, separated by a pooling mechanism, while the discriminator is mainly an encoder. In SGAN, a variety loss is introduced on top of the adversarial (min-max) loss in order to encourage the generator to output diverse samples, thanks to a noise distribution injected to the pooling mechanism output. For details on the SGAN architecture, the reader is advised to refer to the original work~\cite{gupta2018social}. 
%We provide details on the loss functions in the appendix. 
The performance measures used in SGAN for the evaluation process are the absolute displacement error (ADE) and the final displacement error (FDE) of the predicted trajectory ($\Tilde{X}$). The measures are calculated as follows:
\begin{equation}
    ADE = \frac{\sum_{i=1}^N \sum_{t=1}^{T_{pred}} \| \Tilde{X}^i_t - X^i_t\|_2}{N * T_{pred}}
\end{equation}

\begin{equation}
    FDE = \frac{\sum_{i=1}^N  \| \Tilde{X}^i_{T_{pred}} - X^i_{T_{pred}}\|_2}{N}
\end{equation}
where $N$ is the total number of training trajectories.

\begin{figure*}
\centering
\includegraphics[trim={0 7cm 0 3.5cm},clip,scale=0.42]{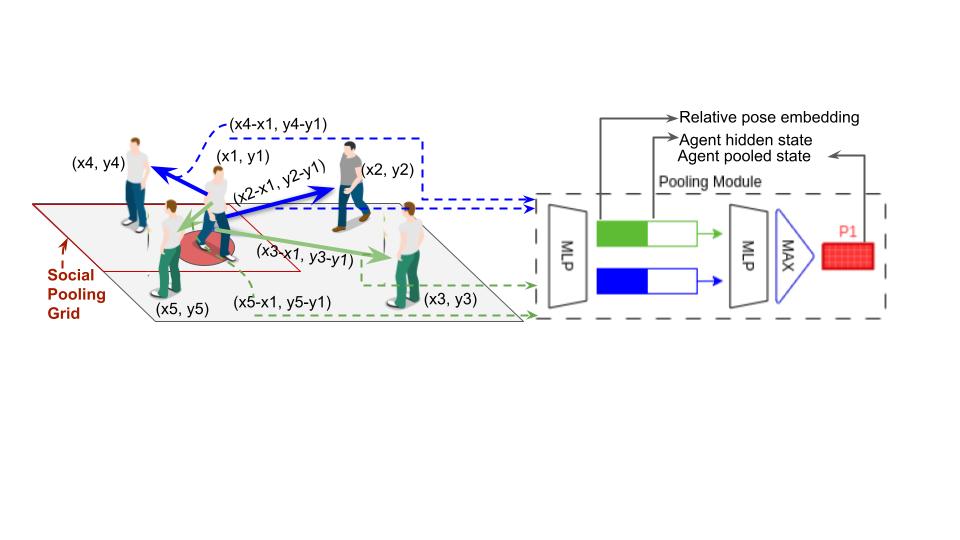}
\caption{The neuro-symbolic SGAN pooling mechanism. The difference with the original SGAN pooling mechanism can be seen from the mixed arrows colour added within and outside the red grid to represent different types of spatial relations or interactions with the central agent standing on the red spot, which can be inferred from the NeuroSyM SGAN architecture.}\label{fig:symsgan}
\end{figure*}

The neuro-symbolic version of SGAN proposed in this paper is illustrated  in Fig.~\ref{fig:symsgan}, highlighting the difference to the original pooling mechanism of SGAN~\cite{gupta2018social}. NeuroSyM acts mainly on the pooling mechanism of the predictive models, where it represents human-human interactions by (a) embedding first their relative pose in all the observed states of each agent through a dense layer, then (b) weighing the embedded relative pose based on the CND-inspired label ($\alpha_{cnd}$) associated to the interaction at a previous time step, and finally (c) max-pooling the weighted embedding across neighbours in the global scene. On the contrary, the original SGAN considers relative poses at the final observed state only, with no attention given to the reliability or stability level the interactions might have to help inferring future states of the agent under consideration.

\textit{\textbf{Results:}}
For a reliable comparison between SGAN and NeuroSyM SGAN, we trained again the former on our computing system (11th Gen Intel® Core™ i7-11800H processor and NVIDIA GeForce RTX 3080 16GB GPU), which was able to replicate almost the same hyper-parameters of the original work on SGAN, except for the batch size, in our case limited to 10 instead of 64. A comprehensive list of the hyper-parameters used to train and validate all model architectures is reported in the appendix Sec.~\ref{sec:app}. The ADE and FDE results for both architectures are reported together with their standard deviations (DE-STD and FDE-STD) in Table~\ref{tab:sgan} for $T_{pred} =$  8 steps (i.e. 3.2~seconds) and 12 steps (i.e. 4.8~seconds), and on the five sequences from the publicly available datasets ETH and UCY.
The results show a better ADE, FDE, DE-STD, and FDE-STD for the NeuroSyM approach compared to the original SGAN. The relative gain in terms of error drop is represented in Table~\ref{tab:sgan} by a positive percentage for all the four measures with NeuroSyM with respect to SGAN on each dataset. The average relative gain for ADE, FDE, DE-STD, and FDE-STD, over the 5 datasets, is 60.84\%, 58.4\%, 28\%, and 33.68\%, respectively, for $T_{pred}=8$; and 78.58\%, 76.97\%, 43.5\%, 46.3\%, respectively, for $T_{pred}=12$. 

\begin{table*}
\begin{center}
\begin{tabular}{p{1.0cm} || p{2.4cm} | p{1.7cm} | p{1.7cm} | p{1.7cm} | p{1.7cm} | p{1.7cm} | p{1.5cm} }
 Measure & Model  & Zara1 &  Zara2 &  Hotel & Univ & ETH & Mean Gain \\ 
\hline 
\hline
\multirow{3}{*}{ADE} & Baseline (SGAN) & 0.7 / 2.29 &   0.44 / 0.95 & 1.76 / 2.45 &  1.25 / 2.96 & 0.88 / 3.8 &  \hspace{15pt}--- \\ 
\cline{2-8}
& NeuroSyM (SGAN) & \textbf{0.21} / \textbf{0.34} &  \textbf{0.2} / \textbf{0.3} &   \textbf{0.35} / \textbf{0.5} & \textbf{0.36} / \textbf{0.62} & \textbf{0.63} / \textbf{0.73} & \hspace{15pt}--- \\
\cline{2-8}
& Relative Gain (\%)  & +70 / +85.15 &  +54.5 / +68.4 & +80.10 / +79.6 &  +71.2 / +79 & +28.4 / +80.78 & +60.84/+78.58\\
\hline
\hline
\multirow{3}{*}{FDE} & Baseline (SGAN) & 1.31 / 4.33 &  0.84 / 1.85 & 3.33 / 4.55 &  2.31 / 5.79 & 1.63 / 6.71 &  \hspace{15pt}---\\ 
\cline{2-8}
& NeuroSyM (SGAN) & \textbf{0.41} / \textbf{0.7} &  \textbf{0.4} / \textbf{0.61} &  \textbf{0.67} / \textbf{0.99} & \textbf{0.74} / \textbf{1.31} & \textbf{1.25} / \textbf{1.44} &  \hspace{15pt}---\\
\cline{2-8}
& Relative Gain (\%) & +68.7 / +83.8 &  +52.38 / +67 & +79.88 / +78.2 &  +67.9 / +77.37 & +23.3 / +78.5 & +58.4/+76.97\\
\hline
\hline
\multirow{3}{*}{DE-STD} & Baseline (SGAN) & 0.35 / 0.9 &  0.26 / 0.52 & 0.44 / 0.9 &   0.51 / 0.84 & 0.37 / 1.02 &  \hspace{15pt}---\\ 
\cline{2-8}
& NeuroSyM (SGAN) &  \textbf{0.22} / \textbf{0.4} &  \textbf{0.2} / \textbf{0.35} &  \textbf{0.32} / \textbf{0.56} & \textbf{0.24} / \textbf{0.38} &  \textbf{0.37} / \textbf{0.64} &  \hspace{15pt}--- \\
\cline{2-8}
& Relative Gain (\%) & +37.14 / +55.5 &  +23 / +32.7 & +27.27 / +37.7 &  +52.9 / +54.7 & +0 / +37.2 & +28/+43.5 \\
\hline
\hline
\multirow{3}{*}{FDE-STD} & Baseline (SGAN) & 1.1 / 2.86 &   0.8 / 1.61 & 1.51 / 2.88 & 1.6 / 2.63  & 1.15 / 3.29 &  \hspace{15pt}---\\ 
\cline{2-8}
& NeuroSyM (SGAN) & \textbf{0.64} / \textbf{1.2}  &  \textbf{0.58} / \textbf{1.05}  &  \textbf{0.96} / \textbf{1.72} & \textbf{0.68} / \textbf{1.14}  & \textbf{1.09} / \textbf{1.91}  &  \hspace{15pt}--- \\
\cline{2-8}
& Relative Gain (\%)  & +41.8 / +58 &  +27.5 / +34.78 & +36.4 / +40.27 &  +57.5 / +56.6 & +5.2 / +41.9 & +33.68/+46.3 \\
\hline
\end{tabular}
\end{center}
\caption{Performance comparison between the baseline architecture SGAN and its neuro-symbolic approach across all datasets. We report results in the format of 8/12 prediction time steps. ADE, FDE, DE-STD, and FDE-STD measures are in meters and in bold is highlighted the better measure among the two approaches. The lower error the better. The mean gain represents the mean of the relative gains over the 5 datasets, hence it only applies to the relative gain rows.} \label{tab:sgan}
\end{table*}

\subsection{Neuro-Symbolic DA-RNN} \label{sec:darnn}

\textit{\textbf{NeuroSyM DA-RNN Architecture:}} The original DA-RNN architecture~\cite{qin2017dual} implements a dual-stage attention mechanism for time-series forecasting. The dual-stage network consists of an encoder with an input attention module weighing the $n^*$ time-series data spatially, each of length $T_h$, where $T_h$ is the observed time history. The encoder is then followed by a decoder with a temporal attention layer, capturing the temporal dependencies in the input series. The encoder and decoder are based on an LSTM recurrent neural network. The network outputs the prediction of one time-series data of length $T_f$, where $T_f$ is the predictive time horizon. The reader can refer to~\cite{qin2017dual} for a detailed explanation of the network components, where $T_f$ was limited to 1.

\begin{figure*}[t]\centering
\includegraphics[scale=0.38]{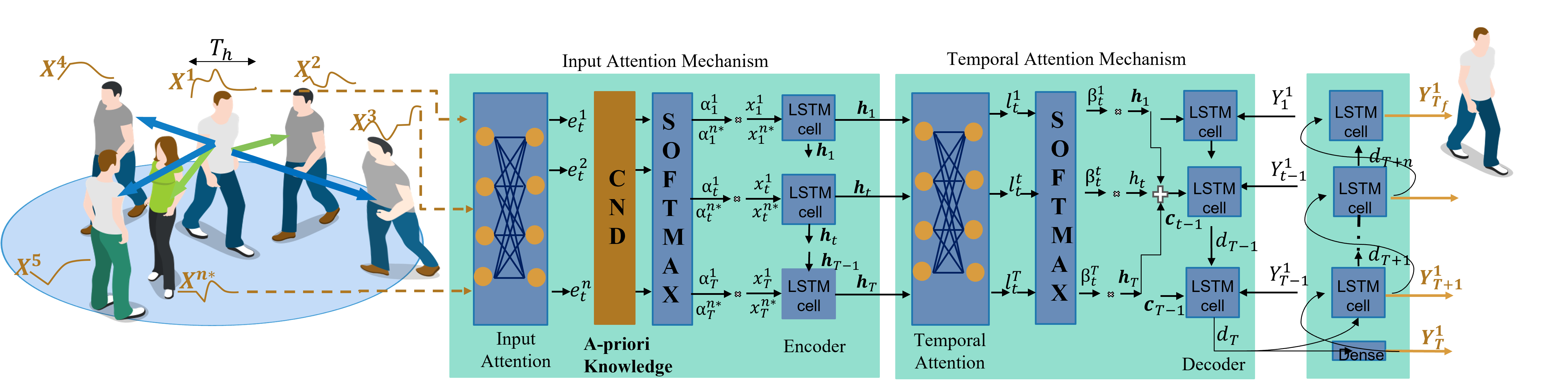}
\caption{A neuro-symbolic approach for attention-based time-series prediction models. Differently from SGAN-like architectures, attention-based mechanisms have no pooling modules. The diagram is extended from~\cite{qin2017dual} and modified for multi-steps attention-based context-aware human motion prediction in crowded environments. The input are $n^*$ time series of agents, within a cluster centered at the first time series, while the output is the prediction of the cluster center's agent. The vector $\mathbf{e}$ denotes the input embeddings normalised to $\alpha$ after passing through the CND layer, which adds a-priori knowledge to them in the form of $\alpha_{cnd,t}^k$. The CND layer weights differently the spatial relations (represented by mixed arrows colour) of the neighbour agents with the central one. The vector $\mathbf{l}$ denotes the temporal attention weights of the encoder's hidden states output, normalised to $\beta$, while $\mathbf{c}$ represents the context. $\mathbf{X}=\{x,y\}$ is the input driving vector; $\mathbf{Y}=\{x', y'\}$ is the label vector; $\mathbf{h}$ and $\mathbf{d}$ are the encoder and decoder hidden states, respectively. The input and temporal attention layers are constructed from dense layers.}\label{fig:network}
\end{figure*}

The NeuroSyM version of DA-RNN we propose in this paper takes advantage of the symbolic knowledge of the spatial interactions between pairs of agents. In DA-RNN, the encoder attention weights (``$\alpha$'' in Fig.~\ref{fig:network}) highlights the importance of each input series at time $t$ on the output prediction at $t+1$. The input attention weights in DA-RNN are calculated as follows:
\begin{equation}\label{eq:soft}
   \alpha_t^k = \frac{\exp(e_t^k)}{\sum_{i=1}^n \exp(e_t^i)}
\end{equation}
where $e_t^k$ is the embedding of the $k^{th}$ input series at time~$t$. It is implemented as:
\begin{equation} \label{eq:embed}
   e_t^k = dense[tanh(dense(\mathbf{h}_{t-1}; \mathbf{s}_{t-1}) + dense(\mathbf{x}^k_{1..T_h}))]
\end{equation}
where $h_{t-1}$ and $s_{t-1}$ are the hidden and cell state of the encoder LSTM at a previous time step. The NeuroSyM DA-RNN acts on the input series embedding $e_t^k$ before the softmax function (Eq.~\ref{eq:soft}) is applied on it. Hence, the NeuroRoSyM approach transforms Eq.~\ref{eq:embed} into $\alpha_{{cnd},t}^k$  $e_t^k$, updating the encoder attention weights with an a-priori knowledge of the reliability or stability of each input series. For applications of human motion prediction in crowds (i.e. with context), $\alpha_{{cnd},t}^k$ is generated from Eq.~\ref{eq:labeling}. Each input series represents the motion history of a neighbour agent, whereas the first time series is the motion history of the considered person and the output is the predicted motion of that specific agent.
Fig.~\ref{fig:network} illustrates schematically where the NeuroSyM module intervenes on the original DA-RNN architecture with the injection of a CND layer at the interface between the embedding and the softmax layers. 

\textbf{Data Processing:} 
Social dense scenarios as the ones presented in the JackRabbot dataset often have an unpredictable number of people entering ($P_e$) and leaving ($P_l$) the environment, possibly leading to a combinatorial explosion in the input size of the predictive model and in its number of training parameters (i.e. when $P_e \gg P_l$ and $P_e$ is very large). Indeed, $n$ individuals in a scene results in $n(n-1)/2$ pairwise data points. That is in addition to the difficulty of deploying an online model with variable input size. 
As a consequence, we implement a crowd clustering approach on JRDB for local interactions embedding (as shown in Figs.~\ref{fig:intro} and~\ref{fig:network}). For each agent $i$ in a given scene, we generate a cluster with a fixed interaction radius $R=3.7m$. The latter is selected based on the proxemics' literature~\cite{hall1969hall}, where the social distance for interactions among acquaintances is indeed between $2.1m$ and $3.7$m. Each cluster includes $n$ input series, with $n$ being the maximum number of agents entering the cluster of agent $i$ in a time interval $T$. The maximum number of input series among all clusters, $n^*$, is fixed for practical (training) purposes. 
 %The number $n^*$ is therefore the maximum size of all the generated clusters. 
 Each cluster is then post-processed to include $(n^* - n)$ input series with complementary ``fake'' values.
 We make use of the open-source annotated 3D point clouds from JRDB, provided as metric coordinates of human (dynamic context) bounding boxes centroid, and extracted from the upper Velodyne sensor, as raw data and ground truth to our network architecture.
The raw data are further processed to extract QTC representations of spatial interactions between pairs of agents in a cluster. We then make use of the CND dictionary to associate each local QTC representation to its corresponding weight $\alpha_{cnd,t}^k$, which is then called by the NeuroSyM architecture as an a-priori information on the reliability of each input series and hence their importance on the predicted output. The a-priori information is then used to weight the embedding of the inputs. The environments considered in JRDB are fairly crowded. Among them, we selected a cafe shop (\textit{bytes-cafe-2019-02-07\_0}).
%Though we omit the integration of static context in this work, the spatial coordinates of the selected key objects (such as bar order and check-out points, exit door, drinking water station, as illustrated in Fig.~\ref{fig:intro}) can be incorporated in the network architecture as any other dynamic agent.
DA-RNN embeds input series using dense layers, facilitating the possibility to integrate static context as series of constant metric coordinates. In the cafe scenario, the static context includes objects such as bar order and check-out points, exit door, and drinking water station, as illustrated in Fig.~\ref{fig:intro}. These objects were manually selected based on a previous investigation to identify the most common ones used by people in the scenario, although in the future we plan to learn them automatically in order to adapt to different environments. The spatial coordinates of the selected key objects are incorporated in the network architecture as any other dynamic agent. \\
\textit{\textbf{Results:}}
We evaluated the DA-RNN and the NeuroSyM DA-RNN for the cafe scenario over to medium (i.e. 48 time steps, or $3.2$ seconds) and long (i.e. 80 time steps, or $5.33$ seconds) term horizons. The parameters for medium and long term horizon prediction were chosen based on relevant literature of human motion prediction~\cite{gupta2018social,tao2020dynamic}. The results of DA-RNN architecture on the JRDB dataset, and the NeuroSyM DA-RNN, are illustrated in Table~\ref{tab:darnnr}, showing root mean square error (RMSE) and mean absolute error (MAE) between the predicted $(x', y')$ coordinates and their true labels $(x, y)$.
We can clearly see that the NeuroSyM version of the architecture succeeds in decreasing the RMSE metric by 22\% on the 48 steps prediction horizon, while influencing the performance negatively by 4\% on the longer 80 steps horizon. At the same time, the neuro-symbolic approach decreased the MAE by 21\% on the 80 steps horizon, while influencing negatively by 3\% the 48 time steps prediction. Although the improvement percentage of NeuroSyM DA-RNN is superior, by a large extent, to its counterpart on one of the two horizon windows, the unequal performance suggests that the prediction is affected by some outliers on the long term. This issue will be addressed in our future work by exploiting the influence of cluster radius selection on the prediction, a factor that we foresee to affect the hidden states embedding of the relative context. For a complete performance evaluation, another fundamental point that we will address in the future is the difference in computational cost between neuro-symbolic architectures and their neural counterparts.

\begin{table}
\begin{center}
\begin{tabular}{p{3.3cm} | p{1.8cm} | p{1.8cm}}
 Architecture &  RMSE & MAE \\ 
\hline 
\hline
%No Input Attention &  \textcolor{red}{2.745 / 3.508} & \textcolor{red}{2.153/ 2.747}\\
%\hline
DA-RNN (Baseline) & 3.61 / \textbf{3.572} & \textbf{2.097} / 2.753  \\
\hline
NeuroSyM DA-RNN & \textbf{2.815} / 3.728 &  2.162 / \textbf{2.166} \\
\hline
Relative Gain (\%) & +22 / -4.37 & -3.1 / +21.32 \\
\hline
\end{tabular}
\end{center}
\caption{Performance comparison between the baseline architecture DA-RNN and the NeuroSyM approach on the JackRabbot dataset. The results' format refers to the 48/80 prediction time steps. RMSE and MAE values are in meters, and the best results are highlighted in bold (i.e. the lower error, the better).} \label{tab:darnnr}
\end{table}

%\begin{figure}\centering
%\includegraphics[trim={2.1cm 0 2.1cm 0},clip, scale=0.39]{dwg/darnn.png}
%\caption{RMSE and MAE measures from testing the DA-RNN architecture without input attention mechanism, with input attention and lastly with input attention updated from a-priori information on spatial interactions in multi-agents scenario.}\label{fig:darnnr}
%\end{figure}

\section{Conclusion} \label{sec:conc}

In this work, we presented a neuro-symbolic approach for context-aware human motion prediction (NeuroSyM) in dense scenarios, leveraging a qualitative representation of interactions between dynamic agents to assess the type of neighbourhood interactions and weight them accordingly. We formulate the spatial interactions in terms of a qualitative trajectory calculus (QTC) and we use the conceptual neighbourhood diagram (CND) to anticipate possible interactions that might influence the future state of an individual agent. The likelihood of the next-step interaction state is used for labeling it given the current state of each agent (position and interaction). We tested the NeuroSyM approach on a fundamental baseline architecture for context-aware human motion prediction, i.e. SGAN, and on another baseline architecture for multivariate time-series prediction, i.e. DA-RNN. Differently from SGAN, where interactions are pooled altogether, DA-RNN includes an attention mechanism on the input time series. We show that, in most of the cases, our neuro-symbolic approach outperforms the baseline architectures in terms of prediction accuracy on the medium- and long-term horizons.
We plan in our future work to test the NeuroSyM approach for motion prediction on other architectures, as S-LSTM and those incorporating static context in addition to the dynamic one. Also, we will exploit the proposed neuro-symbolic approach for human motion prediction in social robot navigation environments, incorporating causal (symbolic) models from literature as~\cite{seitzer2021causal, castri2023causal}. 

\section*{Appendix: Hyper-parameters} \label{sec:app}
The hyper-parameters used to train and validate each of the network architectures deployed in this work are specified in Tab~\ref{tab:hyperparams}. For a complete list of SGAN hyper-parameters and a better understanding of their roles, the reader should refer to the original open-source repository at \url{https://github.com/agrimgupta92/sgan}. 
\begin{table}
    \centering
    \begin{tabular}{ll|ll}
        HP$^{SGAN}$  & Val$^{SGAN}$ & HP$^{DA-RNN}$ & Val$^{DA-RNN}$ \\
        \hline
        Batch size  & 10  & Batch size & 5  \\
        num\_iterations & 8512 & observed\_timesteps  & 5 \\ 
        num\_epochs & 200 & num\_epochs & 80 \\
        noise\_dim & 8 & learning\_rate & 0.001 \\
        noise\_type & Gaussian & train\_ratio & 0.8\\
        pool\_every\_timestep & 0 & decay\_frequency & 1000\\
        observed\_timesteps  & 8 & decay\_rate & 0.99\\
        embedding\_dim & 16 & num\_labels & 1 \\
        encoder\_h\_dim\_g & 32 & encoder\_h\_dim & 256\\
        bottleneck\_dim  & 8   &  decoder\_h\_dim & 256\\
        decoder\_h\_dim\_g & 32 & validation\_ratio & 0.1\\
        encoder\_h\_dim\_d & 48 & test\_ratio & 0.1\\
        g\_learning\_rate & 0.0001 \\
        d\_learning\_rate & 0.001 \\
        mlp\_dim & 64 \\
        noise\_mix\_type & global \\
        \hline
    \end{tabular}
    \caption{(Left) SGAN and NeuRoSyM SGAN hyper-parameter values. (Right) DA-RNN and NeuRoSyM DA-RNN hyper-parameter values.}
    \label{tab:hyperparams}
\end{table}

\bibliographystyle{IEEEtran}
\bibliography{IEEEabrv,ijcnn23}

\end{document}